# A Novel Efficient Task-Assign Route Planning Method for AUV Guidance in a Dynamic Cluttered Environment

S. MahmoudZadeh, D. M.W. Powers, A.M. Yazdani

School of Computer Science, Engineering and Mathematics
Flinders University, Adelaide, SA 5042, Australia

*Abstract*— **Promoting the levels of autonomy facilitates the vehicle in performing long-range operations with minimum supervision. The capability of Autonomous Underwater Vehicles (AUVs) to fulfill the mission objectives is directly influenced by route planning and task assignment system performance. This paper proposes an efficient task-assign route planning model in a semi-dynamic operation network, where the location of some waypoints are changed by time in a bounded area. Two popular meta-heuristic algorithms named biogeography-based optimization (BBO) and particle swarm optimization (PSO) are adopted to provide real-time optimal solutions for task sequence selection and mission time management. To examine the performance of the method in a context of mission productivity, mission time management and vehicle safety, a series of Monte Carlo simulation trials are undertaken. The results of simulations declare that the proposed method is reliable and robust particularly in dealing with uncertainties and changes of the operation network topology; as a result, it can significantly enhance the level of vehicle's autonomy by relying on its reactive nature and capability of providing fast feasible solutions.**

## I. INTRODUCTION

Autonomous Underwater Vehicles (AUVs) are advantageous tools for undersea exploration, interrogation, detection and surveillance and particularly are employed to accomplish tasks that are impossible for human operator to complete. Most of the current AUV applications are supervised from the support vessel which provides higher-level decisions in critical situation and generally takes enormous cost during the mission [1]. Growing attention has been devoted in recent years on increasing the ranges of missions, vehicles endurance, extending vehicles applicability, promoting vehicles autonomy to handle longer missions without supervision, and reducing operation costs [2]. The primary step toward increasing endurance and range of vehicle operation is promoting vehicles autonomy in terms of time management and task allocation while moving toward the destination. More advanced approaches thus aim to increase the efficiency of the vehicle in both robust decision-making and situation awareness. Efficient motion planning and mission scheduling are also key requirements towards advanced autonomy, and facilitate the vehicle's handling of long-range operations.

Route planning problem usually refers to finding shortest paths in a graph-like network such as modelling the transportation network [3, 4]. The main issue addressed by previous research on route planning system is how to direct vehicle(s) to destination(s) in a network while providing efficient maneuvers and reducing travel time. Some instances of route planning systems applications are in the areas of traffic control [5], real time routing and trip planning [6], and so on. Briefly reviewing the most highlights in route planning works in the state of the art, in [7] a route planning strategy is employed for transportation purpose in a form of multi-agent decisions in which the agent is in charge of order distribution to customers, traversing edges, competing vendors, increasing production and etc.; a three-layer structure to facilitate multiple unmanned surface vehicles to accomplish task management and formation path planning in a maritime environment is proposed in [8]; in [9], graph-based methods using modified Dijkstra Algorithm for the AUV ''SLOCUM Glider'' motion planning in a dynamic environment is offered; for AUV guidance in large scale underwater environment, a behavior based controller coupled with waypoint tracking scheme is employed [10]; and finally, a special model of multi-agent reinforcement learning (MARL) algorithms is proposed in [11] for a road network route planning system taking advantages of Q-value based dynamic programming (QVDP) to solve vehicle delay's problems .

With respect to the combinatorial nature of AUV's route-task allocation problem, which generalizes both TSP and Knapsack problems, there should be a compromise among the mission available time, maximizing number of highest priority tasks with minimum risk percentage, and guaranteeing reaching to the predefined destination, which is combination of a discrete and a continuous optimization problem at the same time and categorized as a Non-deterministic Polynomial-time (NP) hard problem. Obtaining the optimal solutions for NP-hard problems is a computationally challenging issue and currently it is impossible to find a polynomial time algorithm that solves an NP-hard problem of even moderate size. Moreover, obtaining the optimum solution is only possible for the particular case where the environment is completely known and no uncertainty exists. However, the modelled environment in the proceeding research corresponds to a dynamic network with high uncertainty.

The problem size and complexity grows exponentially with increasing the size of the operation network (number of nodes/connections). Therefore, handling the complexity of the network topology or in general problem space results in a demanding computational burden, which is an intricate issue to be considered. Meta–heuristics algorithms are the fastest methods introduced for solving NP-hard complexity of vehicle routing problem and tend to produce near optimal solutions with high probability [12]. Although the captured solutions by any meta–heuristic algorithm do not necessarily correspond to the optimal solution, it is more important to control the computational time to cover real-time performance of the AUV route planning. Thus, relying on the previously mentioned ability of meta-heuristic algorithms, the BBO and PSO algorithms are employed to find correct and near optimal solutions in competitive CPU time.

As AUVs operate in an uncertain environment, there is a huge amount of variability in the travel times, which can have a devastating effect on mission plans. Unlike previous research on vehicle routing problems, which mostly look for the shortest possible route in a graph, this research aims to complete the maximum number of tasks for a model in which time and distance are a function of the individual task. Several cost factors such as route length, travel time, task priority and task specific metrics must be simultaneously minimized or maximized in order to make maximum use of the available time but not exceeding it, rather than just looking for a shortest route. Proper time management of the vehicle routing operations is necessary to ensure on-time mission completion and consequently the mission success. This paper is an extension of [13], in which the PSO and GA algorithms have been adopted to solve the vehicles routing problem in a static operation network assuming that the position of waypoints are known in advance where no offline map data is encountered. In the proceeding research, similarly the priority based approach for valid route generation is applied and an efficient BBO and PSO based routing strategy is provided to handle task assignment and time management in a semi-dynamic operation network where the vehicle experiences both fixed waypoints and moving ones in a bounded zone due to ocean current force. The dynamic nodes are assumed as the wireless sensors that used to update AUV's knowledge about environmental changes. The AUV can update its mission by reaching to one of these nodes and re-routing may be required when remarkable deformation is appeared in the network topology (when the position of the dynamic nodes updated). A real map data is conducted and clustered by k-means method to model a realistic marine environment. In the proceeding paper, the task assign-routing problem is detailed in section II. An overview of the PSO and BBO algorithms is provided in section III and section IV, respectively. Application of the BBO and PSO algorithms on the stated problem is demonstrated by section V. The simulation results is discussed in section VI, and section VII concludes the paper.

## II. MATHEMATICAL PRESENTATION OF TASK ASSIGN-ROUTING PROBLEM

Route planning aims to find reasonable route among several waypoints that starts from a particular point and should reach to the destination point after meeting adequate number of waypoints. Apparently it is impossible for a single vehicle to cover all tasks in a single mission in a large-scale operation area. The applicable variables in graph routing problem (e.g. priority and risk of assigned tasks to network edges) can be used to provide a priority tour and beneficial mission for vehicle. Reaching to the destination is the second critical issue for a route planner that should be taken into consideration. Hence, existing tasks that assigned to graph edges are selected and prioritized in a way that govern the vehicle to the destination.

Existence of prior information about the terrain, location of coasts as the forbidden zones for deployment, and position waypoints in operating area promotes AUV's capability in robust motion planning. To model a realistic marine environment, a three dimensional terrain $\{\Gamma_{3D}\}$ is considered based on real map of the Benoit's Cove (located in Newfoundland, Canada [14]), in which the terrain is covered by fixed and uncertain dynamic waypoints. To this purpose, a map in size of 500×1000 pixels is used that corresponds to area of 5×10 $km^2$, where each pixel represents 10 $m^2$. A k-means clustering method is applied to cluster water zone as valid sections for deployment. Waypoints are located in the joint water covered zone.

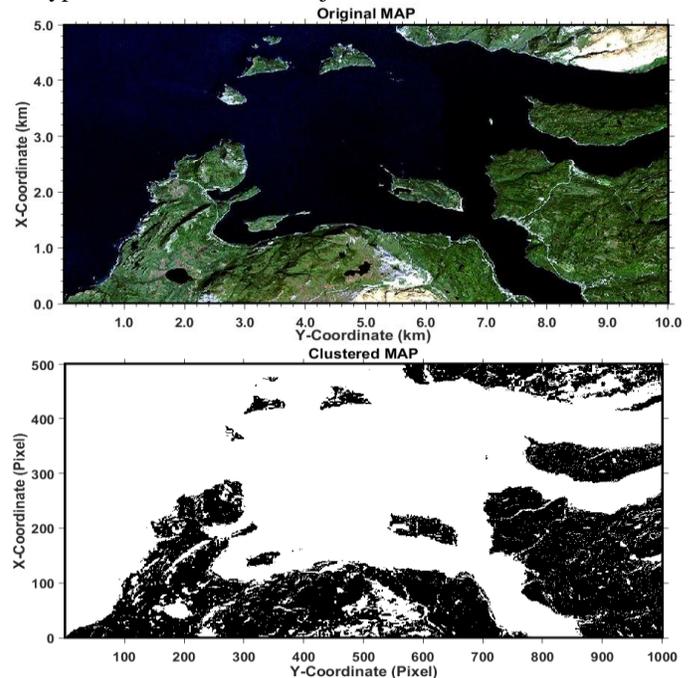

Fig. 1. (a) The original map of the Benoit's Cove [14]. (b) The clustered map in which the area is clustered to forbidden (black) and valid (white) zones for AUV's deployment.

Existing waypoints are divided into two categories as follows:

- The static waypoints $SP_{x,y,z}$ with known positions initialized once in advance with uniform distribution of ~$U(0,5000)$ for $SP^i_x$, ~$U(0,10000)$ for $SP^i_y$ and ~$U(0,1000)$ for $SP^i_z$.

- Dynamic waypoints $DP_{x,y,z}$ considered as underwater wireless sensors that used to transfer local information to the vehicle. The location of these waypoints is sensed by the sonar sensors with a specific uncertainty modelled with a normal distribution of ~$N(0,\sigma^2)$ and gets updated within specified area during vehicles deployment.

$$\begin{aligned} Bound^{\max}_{x,y,z} &\sim N(0,\sigma^2) \\ \left| DP^i_{x,y,z} \right| &\leq Bound^{\max}_{x,y,z} \end{aligned} \qquad (1)$$

Therefore the waypoints position has a truncated normal distribution, where its probability density function defined as follows:

$$\begin{aligned} f\left(DP^i;0,\sigma^2,Bound^{\max}_{x,y,z}\right) &= DP^i_{x,y,z} \Big/ \left|2 \times Bound^{\max}_{x,y,z}\right| \\ DP_{x,y,z} & \& DP^i_{x,y,z} \notin \{MAP=0\} \end{aligned} \qquad (2)$$

Various tasks assigned to passible distance between connected waypoints in advance. Hence, each edge is weighted by a function of tasks priority, risk percentage, and tasks completion time. The route cost is determined based on connections length, weight, and time required for traversing edges included in the route.

$$\begin{aligned} &\forall DP^i_{x,y,z} \& SP^i_{x,y,z} \notin \{MAP=0\}; \ \{P_{x,y,z}\} = \{SP_{x,y,z}\} \cup \{DP_{x,y,z}\} \\ &\Re_k = \left(P^1_{x,y,z},...,P^i_{x,y,z},...,P^n_{x,y,z}\right) \\ &\forall DP^i_{x,y,z} \& SP^i_{x,y,z}: \ \exists q_{ij}:(d_{ij}(t),T_{ij}) \end{aligned}$$

$$T_{ij} = \left( \frac{\sqrt{\left(P^i_x(t)-P^i_x(t)\right)^2+\left(P^j_y(t)-P^i_y(t)\right)^2+\left(P^j_z(t)-P^i_z(t)\right)^2}}{|v_{AUV}|} \right) + \delta_{ij}$$

$$\forall q_{ij}: \ \exists Task_{ij}:(\rho_{ij},\xi_{ij},\delta_{ij}) \qquad (3)$$

where, $P^i_{x,y,z}$ represents the coordinate of any arbitrary waypoint in geographical frame, the $v_{AUV}$ is the absolute velocity of the vehicle, $\Re$ is an arbitrary route, $T_{ij}$ represents the required time for traversing the distance $d_{ij}$ between $P^i$ and $P^j$ that is updated iteratively. Location of dynamic waypoints and consequently the length of connections get altered simultaneously; thus, considering this issue is necessary in cost computation and route re-planning process. Each edge in the graph involves the corresponding task's completion time $\delta_{ij}$, priority $\rho_{ij}$ and risk percentage $\xi_{ij}$. The total weight of route $W_{\Re}$ should be maximized and the route travel time should approach available time.

$$\begin{aligned} W_{\Re} &= \sum_{\substack{i=0 \\ j \neq i}}^{n} \left(\frac{lq_{ij}\rho_{ij}}{\xi_{ij}}\right) = \sum_{\substack{i=0 \\ j \neq i}}^{n} \left(lq_{ij} \times w_{ij}\right); \quad l \in \{0,1\} \\ T_{\Re} &= \sum_{\substack{i=0 \\ j \neq i}}^{n} lq_{ij}\left(\delta_{ij} + \frac{d_{ij}(t)}{|v_{AUV}|}\right) \end{aligned} \qquad (4)$$

where $T_{\Re}$ is the required time to pass the route, $l$ is the selection variable for arbitrary edge of $q_{ij}$.

## III. OVERVIEW OF PARTICLE SWARM OPTIMIZATION

The Particle swarm optimization (PSO) is one of the fastest optimization methods employed for solving diverse complex problems since past decades. The process of PSO is initialized with a population of particles. Each particle involves a position and velocity in the search space that get updated iteratively. Each particle preserves its previous state, the best position in its experience $\chi^{P-bst}$ and the global best position of $\chi^{G-bst}$. The cost of particle current position is compared to the $\chi^{P-bst}$ and $\chi^{G-bst}$ at each iteration. More detail about the algorithm can be found in [15]. Particle position and velocity gets updated using (5).

$$\begin{aligned} v_{ij}(t+1) &= \omega v_{ij}(t) + c_1 r_1[\chi^{P-bst}_{ij}(t)-\chi_{ij}(t)] + c_2 r_2[\chi^{G-bst}_{ij}(t)-\chi_{ij}(t)] \\ \chi_{ij}(t+1) &= \chi_{ij}(t) + v_{ij}(t+1) \end{aligned} \qquad (5)$$

where $c_1$ and $c_2$ are acceleration coefficients, $\chi_i$ and $v_i$ are particle position and velocity at iteration $t$. $r_1$ and $r_2$ are two independent random numbers in [0,1]. $\omega$ exposes the inertia weight and balances the PSO algorithm between the local and global search. The argument for using the PSO on route planning problem is strong enough due to its superior capability in scaling well with complex and multi-objective problems. However, PSO has problem in particle coding step due to discrete nature of the search space in vehicle's task assign-routing problem. This issue is resolved using a priority/adjacency based route generation strategy (more detail is given in [13]). The process of PSO-based route planning is given by a flowchart in Fig.2.

## IV. OVERVIEW OF BIOGEOGRAPHY-BASED OPTIMIZATION

The BBO is an evolutionary optimization technique developed based on the equilibrium theory of island biogeography concept [16]. The basic idea of the algorithm inspired by the immigration, emigration, and rate of change in the number of species in an island. Each candidate solution in BBO has a quantitative performance index representing the fitness of the solution called habitat suitability index (HSI). Habitability is related to some qualitative factors known as Suitability Index Variables (SIVs), which is a randomly initialized vector. Therefore, each particular candidate solution has a design variable of SIV, emigration rate of μ and immigration rate of λ. The immigration rate λ is used to probabilistically modify the SIV of a selected solution $h_i$. Then emigration rate (μ) of the other solutions is considered and one of them probabilistically selected to migrate its SIV to solution $h_i$ that is known as migration in BBO. Each given solution $h_i$ is modified according to probability of existence of the $S$ species at time $t$ in habitat $h_i$ that gets updated iteratively by

$$p_s(t+1) = p_s(t)(1-\lambda_s(t)-\mu_s(t)) + p_{s-1}\lambda_{s-1}(t) + p_{s+1}\mu_{s+1}(t) \qquad (6)$$

$$\lambda_s = I*\left(1-\frac{S}{S_{\max}}\right); \quad \mu_s = E*\left(\frac{S}{S_{\max}}\right) \xrightarrow{if \ E=I} \lambda_s + \mu_s = E \qquad (7)$$

where $I$ and $E$ are the maximum immigration and emigration rates, respectively. As habitat suitability improves, the number of its species and emigration increases, and the immigration rate decreases. Very high and very low HIS solutions are not probable, where solutions with medium HIS are comparatively probable. Mutation is required for solution with low probability, while solution with high probability is less likely to mutate. Mutation operator increases the diversity of the population and propels the individuals toward global optima. Hence, the mutation rate $m(S)$ is calculated by

$$m(S) = m_{\max}\left[\frac{1 - p_s}{p_{\max}}\right] \qquad (8)$$

where $m_{max}$ is the maximum mutation rate defined by user, $p_{max}$ is probability the habitat with maximum number of species $S_{max}$. A general overview of BBO mechanism on dynamic route planning is provided in a flowchart given by Fig.3.

## V. BBO AND PSO ALGORITHMS ON DYNAMIC TASK-ASSIGN ROUTE PLANNING APPROACH

With respect to formulated graph-like terrain, the global route planner tends to find the best fitted route to the available time, involving the best sequence of waypoints. AUV starts its mission from initial point of $P^1_{x,y,z}$ and should pass sufficient number of waypoints to reach the destination at $P^n_{x,y,z}$. To this purpose, the global route planner simultaneously tends to determine the efficient route in network, trade-off between prioritizing the available tasks and managing the mission available time. In this context, the proposed task-assign-routing problem can be modelled as a multi-objective optimization problem.

### Particle/Habitat Encoding (Route Generation)

The initial step is generating feasible primary routes as initial population for both PSO and BBO optimization process. Developing a suitable coding scheme for individual representation is the most critical step in implementing BBO and PSO frameworks that has direct impact on performance of the algorithm and optimality of the solutions. Habitats in the proposed BBO correspond to feasible routes as a sequence of nodes while in PSO the feasible routes are encoded via particles. According to prior information of tasks and terrain, feasible routes should be generated, in which the route vectors take variable length, but limited to maximum number of nodes included in the graph. The route should be feasible according to following criteria: a valid route should be commenced and ended with predefined start and target waypoints; it should not include edges that are not presented in the graph; it should not traverse an edge for more than once; the route travel time should not exceed the maximum range of total available time.

A priority-based strategy is conducted by this paper to generate feasible routes, in which a randomly initialized priority vector is assigned to sequence of nodes in the graph. Adjacency information of the graph and provided priority vector get used to add proper node to the route sequence. Adjacency information get updated any time that wireless sensors change their locations. The priority vector for corresponding waypoints takes positive or negative values in the specified range of [-200,100]. Afterward, index of waypoints are added to the route sequence one by one according to priority vector and graph adjacency relations.

For further information refer to [13]. Visited nodes in a route get a large negative priority value that prevents multiple visits to that node. Traversed edges of the graph get eliminated from the adjacency matrix; hence, the selected edge will not be a candidate for future selection in a specific route. This issue reduces the time and memory consumption for routing in large graphs. After individual population is initialized the algorithm start its process of finding best fitted route according to addressed objectives in this research.

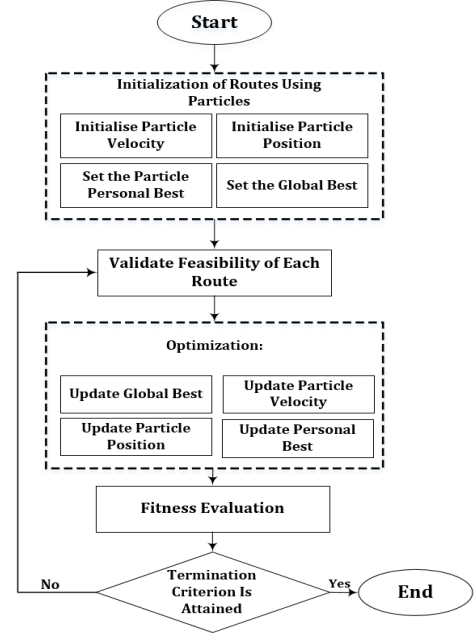

Fig. 2. The process of PSO algorithm on route planning problem

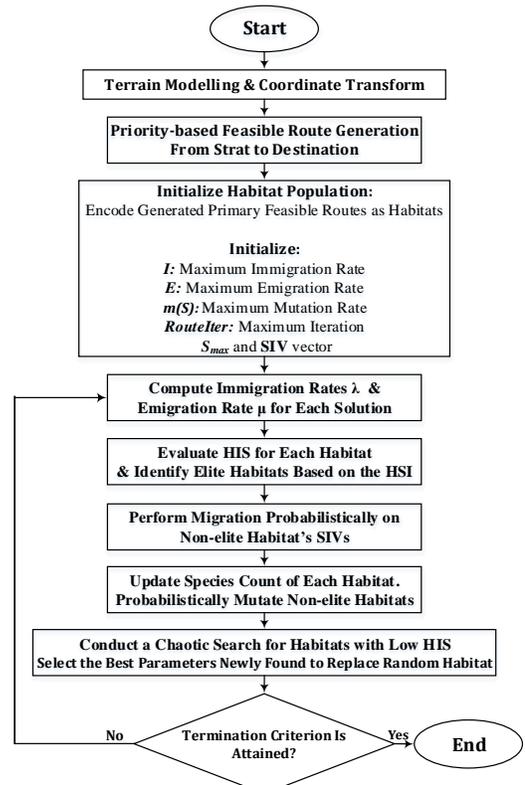

Fig. 3. The process of BBO algorithm on route planning problem

### Route Optimization Criterion

After the individual population is initialized, the optimization process tends to find the best fitted route through the given operation graph. Maximizing highest priority tasks with smallest risk percentage in the time interval that battery's capacity allows, is the main goal of the global route planner. For this purpose, the objective function is defined as a combination of multiple weighted cost functions that should maximized or minimized, indicated by

$$Cost_{\Re} \propto |T_{\Re} - T_{Available}|$$

$$Cost_{\Re} = \varphi_1 \left| \sum_{\substack{i=0 \\ j \neq i}}^{n} lq_{ij} \left( \delta_{ij} + \frac{d_{ij}(t)}{|v_{AUV}|} \right) - T_{Available} \right| + \varphi_2 \sum_{\substack{i=0 \\ j \neq i}}^{n} lq_{ij} \left( \frac{\xi_{\Re}}{\rho_{\Re}} \right) \quad (9)$$

$$s.t.$$
$$\forall \Re_k; \quad \max(T_{\Re_k}) < T_{Available}$$

where $T_{Available}$ is the total available time. The route cost has direct relation to the passing distance among each pair of selected waypoints. As discussed earlier, the locations of dynamic waypoints are changed randomly in the defined boundary; hence, the distances between waypoints $d_{ij}$ is defined based on time and gets updated iteratively during vehicle's deployment. $\varphi_1$ and $\varphi_2$ are two positive coefficients determine amount of participation of the performance factors in the route cost computation. Appropriate setting of these coefficients has a significant effect on performance of the model. After visiting each waypoint in the global route, the re-planning criteria is investigated.

## VI. Discussion on Simulation Results

A configurable dynamic route planner is developed in order to find the most productive optimum route between start and destination points, take maximum use of mission available time, and terminate the mission before vehicle runs out of battery/time. As discussed earlier, some of the waypoints are considered to be floating and their location is variable in a specific boundary, while the rest of the waypoints are fixed and known in position. In this study, it is assumed that tasks are assigned to edges of the graph and the graph parameters initialized once in advance. To evaluate efficiency of the proposed method for a single vehicle routing problem, its performance in task allocation, time management, mission productivity, real-time performance, etc., are tested using BBO and PSO. The argument for using the PSO in solving NP-hard problems such as knapsack or TSP problems is strong enough due to its superior capability in scaling well with complex and multi-objective problems. The PSO is well adapted to multidimensional space and nonlinear functions due to its stochastic optimization nature that does not require any evolutionary operators to transmute the individuals. However, PSO operates in a continuous space originally; hence, a particular problem arises with proper coding of the particles as route candidate due to discrete nature of search space in vehicle's task-assign-routing problem. To handle this shortcoming, a priority based route generation approach has been conducted, which accurately cover this issue but increases the computational burden for this algorithm. Even adding this issue into account, it is clear from Fig.4 that the PSO still shows superior real-time performance comparing to BBO. The BBO also shares some common features with PSO, in which solutions of one generation get transferred to the next. A special feature of the BBO algorithm is that the original population never get discarded but get modified by migration, which this issue promote the exploitation ability of the algorithm. The BBO also uses a mutation operator to increase the diversity of the population that propel the individuals toward global optima. Another specific feature of BBO is that, it uses the fitness of each solution iteratively (for each generation) to assess its emigration/immigration rate and it also has a more effective memory capability comparing the PSO. The algorithms are configured as follows: the maximum number of iterations is set on 150. The PSO optimization configuration is set by 150 particles (candidate paths). The expansion-contraction coefficients also are fixed on *2.0* and *2.5*. The inertia weight decreases from 1.4 to 0.5. For the BBO, The habitats population is set on 50. The number of kept habitats is set on 10. The emigration rate μ is generated in a form of a vector in range of (0, 1), and the immigration rate defined as λ=1- μ. The maximum mutation rate is set on 0.1.

Several performance metrics are employed to assess the functionality of the route planners such as the number of completed tasks, total obtained weight, total cost, total route time, total traveled distance, computational time and the time constraint satisfaction of the generated route with respect to the complexity of the operation network. To compare applied heuristic methods and to evaluate the stability and reliability of the employed algorithms in satisfying performance indexes, 200 execution runs are performed in a Monte Carlo simulation, presented by Fig.4 to Fig.7. The number of waypoints is fixed on 40 nodes for all Monte Carlo runs. Network topology set to change randomly with a Gaussian distribution on the problem search space. The time threshold ($T_{Available}$) also is fixed on $3.1 \times 10^4 (sec)$.

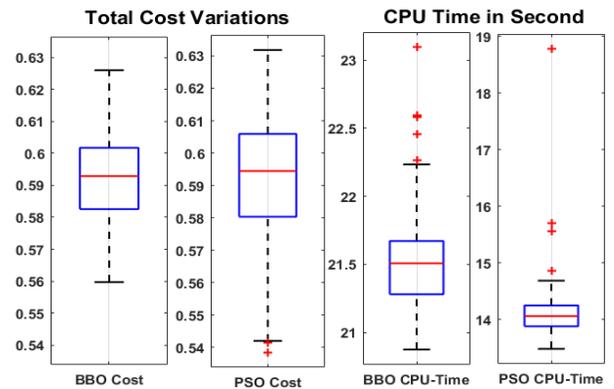

Fig. 4. Average cost and computational time variation for both BBO and PSO on 200 Monte Carlo runs

Fig.4 compares the functionality of BBO and PSO in terms of cost and CPU time variation in dealing with problem's space deformation in several Monte Carlo runs. It is inferred from this comparison that the route cost is varying in a similar range for both BBO and PSO, almost between 0.58 to 0.6. However, the PSO operates faster in producing similar cost. Of course, computational time is a critical factor in such a real-time application; hence faster operation is a significant advantage for PSO-based route planner.

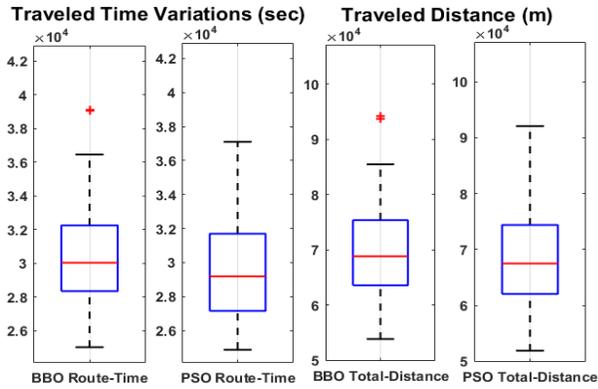

Fig. 5. Variation of route traveled time and distance for both BBO and PSO on 200 Monte Carlo runs

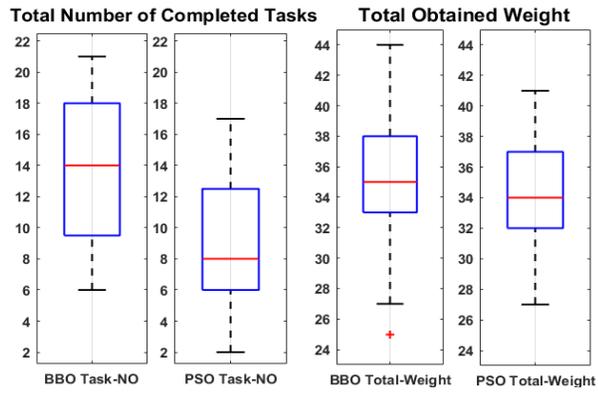

Fig. 6. Average variation of number of completed tasks and obtained weights for both BBO and PSO on 200 Monte Carlo runs

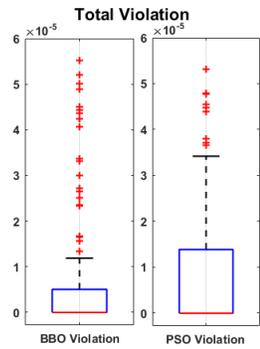

Fig. 7. Average variation of route time violation for both BBO and PSO on 200 Monte Carlo runs

The main purpose of the route planner is taking maximum use of available time (time threshold of $3.1 \times 10^4$ (sec)) but not exceeding that. Considering simulation results in Fig.5, it is clear that both of the BBO and PSO based planners significantly manage the route time to approach proposed time threshold and propose almost similar performance in quantitative measurement of two performance metrics of travel time and total traveled distance.

The provided results in Fig.6 also confirms the superior performance of the PSO based route planner in terms of increasing mission productivity by maximizing total obtained weight and number of covered tasks by taking almost the same traveling time (in Fig.5). From simulation results in Fig.7, it is noted that average variation of route violation for Monte Carlo executions of BBO and PSO based planners approaches zero, which confirms feasibility of the produced route; however, BBO acts more efficiently in converging the solutions into a feasible space and satisfying the defined constraints.

Comparing the capability of BBO and PSO in updating the route, it is noteworthy to mention that each time the position of the dynamic waypoints is changing, the adjacency relation and distance between nodes is updated. In this context, probably the previous route loses its optimality; hence, re-planning a new route would be necessary.

Fig.8 presents an example of such a situation, in which both routes produced by BBO and PSO get updated. The coastal areas are impassable and forbidden for vehicles deployment in Fig.8. The grey circles are the bounds of variation for each dynamic waypoint that initialised in advance with a normal distribution indicating a confidence of 98% that the node is located within this area applying equations given in section II.

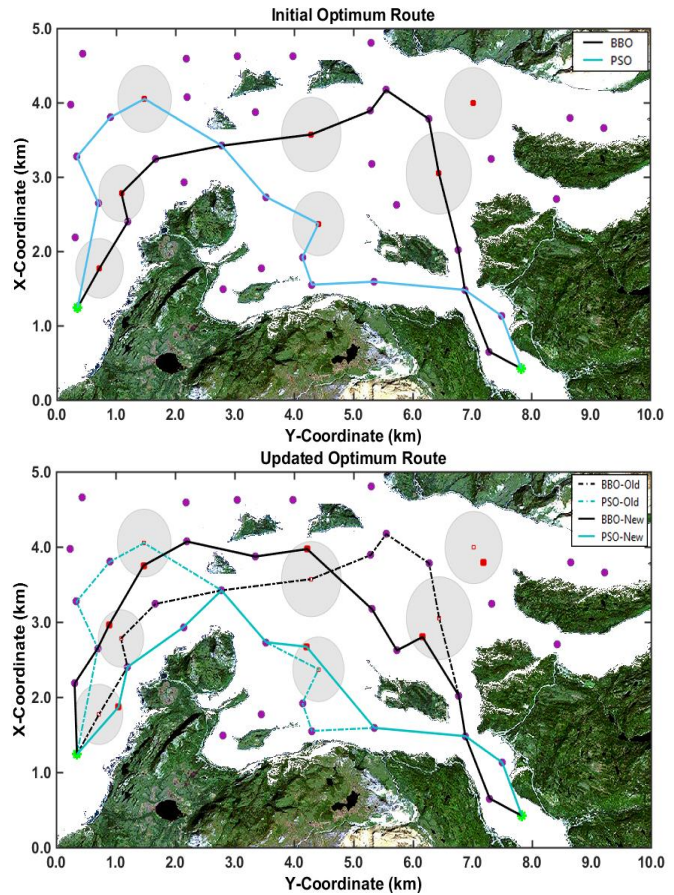

Fig. 8. Route replanning based on network updates and defined constraint.

The route planner on this research simultaneously tracks the measurements of the terrain status. As given in Fig.8, both of the proposed planners are capable of re-generating the alternative trajectory according to latest terrain and both of them are accurate against immediate update of operation network as there is only a slight difference comparing the new and old produced results by both algorithms. The useful information from the previous route is taken into the account for re-planning the route.

As indicated in Fig.4 to Fig.8, both algorithms reveal robust behavior against the variations and meet the specified constraint; however, PSO has superior performance and shows more consistency in its distribution comparing to the generated solutions by BBO algorithm. Indeed it is evident that the performance of both algorithms is relatively independent of network variations and complexity that make them suitable for real-time applications.

## VII. Cossnclusion

This paper investigated the performance of BBO and PSO in providing time optimal routes in a semi-dynamic operation network, while carrying out the mission goals under specific constraints in different graph topologies. Both fixed and moving waypoints, representing specific tasks, were exploited in configuring the problem space. The proposed task-assign-routing problem was encoded in BBO habitats and PSO particles and then using their heuristic search capability the solution was obtained. The simulation results confirmed that the utilized method is capable of generating optimal or near-optimal route not only in a static environment but also in an uncertain semi-dynamic operating field. Additionally, the employed method is computationally fast suitable for real-time applications. The results obtained from Monte Carlo analyses indicated the inherent robustness of both utilized BBO and PSO based route planning in dealing with random configuration of problem space and uncertainty of undersea environment. Future work will focus on modelling a more realistic ocean environment comprising ocean dynamics and involving the kinematics of AUV in route planning configuration.